\documentclass{article}

\usepackage[preprint]{neurips_2026}

\usepackage[utf8]{inputenc} 
\usepackage[T1]{fontenc}    
\usepackage{hyperref}       
\usepackage{url}            
\usepackage{booktabs}       
\usepackage{amsfonts}       
\usepackage{nicefrac}       
\usepackage{microtype}      
\usepackage{amsmath}
\usepackage{color}
\usepackage{graphicx}

\usepackage{algorithm}    
\usepackage{algpseudocode}
\usepackage{wrapfig}      
\usepackage{caption}      

\title{Diffusion Distillation for Efficient Weather Ensembles}

\author{%
  Yiming Yang\thanks{Corresponding to Yiming Yang: \texttt{zcahyy1@ucl.ac.uk}.} \\
  Department of Statistical Science\\
  University College London\\
  \And
  Valentin Brekke \\
  Department of Statistical Science\\
  University College London
  \And
  James Briant \\
  MPLS Doctoral Training Centre\\
  University of Oxford
  \And
  Serge Guillas \\
  Department of Statistical Science\\
  University College London
}

\begin{document}

\maketitle

\begin{abstract}
Diffusion models generate skillful weather ensembles but require costly iterative sampling. We introduce a supervised energy-distance distillation method that compresses a multi-step diffusion teacher into a single-step student by aligning student forecasts with teacher samples and ground-truth observations. Experiments on global forecasting and typhoon-track prediction show that our student outperforms existing distillation methods and preserves skill for extreme events. It matches or surpasses the teacher across key metrics using only one neural function evaluation per autoregressive step. Our code is available at \url{https://github.com/yyimingucl/gencast_distillation}.
\end{abstract}

\section{Introduction}

Machine learning has transformed weather forecasting, achieving strong performance at a much lower cost than numerical weather prediction. Early breakthroughs focused on deterministic forecasting \cite{chen2023fuxi,bi2023accurate,lam2023learning,kurth2023fourcastnet,bodnar2024aurora}, while recent work has shifted toward probabilistic approaches. Diffusion models \cite{ho2020denoising,song2021scorebased} have shown strong probabilistic forecast skill \cite{li2024generative,alet2025skillful,price2025probabilistic}, but their iterative sampling requires many neural function evaluations (NFEs), limiting large ensembles and real-time deployment.

We study diffusion distillation for efficient probabilistic weather forecasting, aiming to compress a multi-step teacher into a one-step student while preserving its predictive distribution. Existing methods broadly follow trajectory matching, which reproduces the teacher’s reverse process \citep{salimans2022progressive,meng2023distillation,DBLP:conf/icml/SongD0S23}, or distribution matching, which aligns their final distributions \citep{yin2024one,yin2024improved}. The former provides indirect supervision of final forecasts, whereas the latter often requires auxiliary networks and GAN-style training. In our experiments, representative methods from both classes produce underdispersed ensembles with lower probabilistic forecast skill than the teacher (Section~\ref{section:Experiment}). A detailed review of related work appears in Appendix \ref{sec: related_work}.

To address these limitations, we propose a supervised distillation framework that aligns the student distribution with the teacher distribution while incorporating ground-truth supervision. Our sample-based distribution-matching objective avoids auxiliary networks and GAN-style training. We apply it to \textsc{GenCast} \citep{price2025probabilistic}, compressing its iterative sampler into a single-pass ensemble generator. Experiments on global forecasting and tropical-cyclone tracking show that our student outperforms existing distillation methods and preserves the teacher’s forecast skill at a substantially lower inference cost.

\begin{figure}
    \centering
    \includegraphics[width=0.95\linewidth]{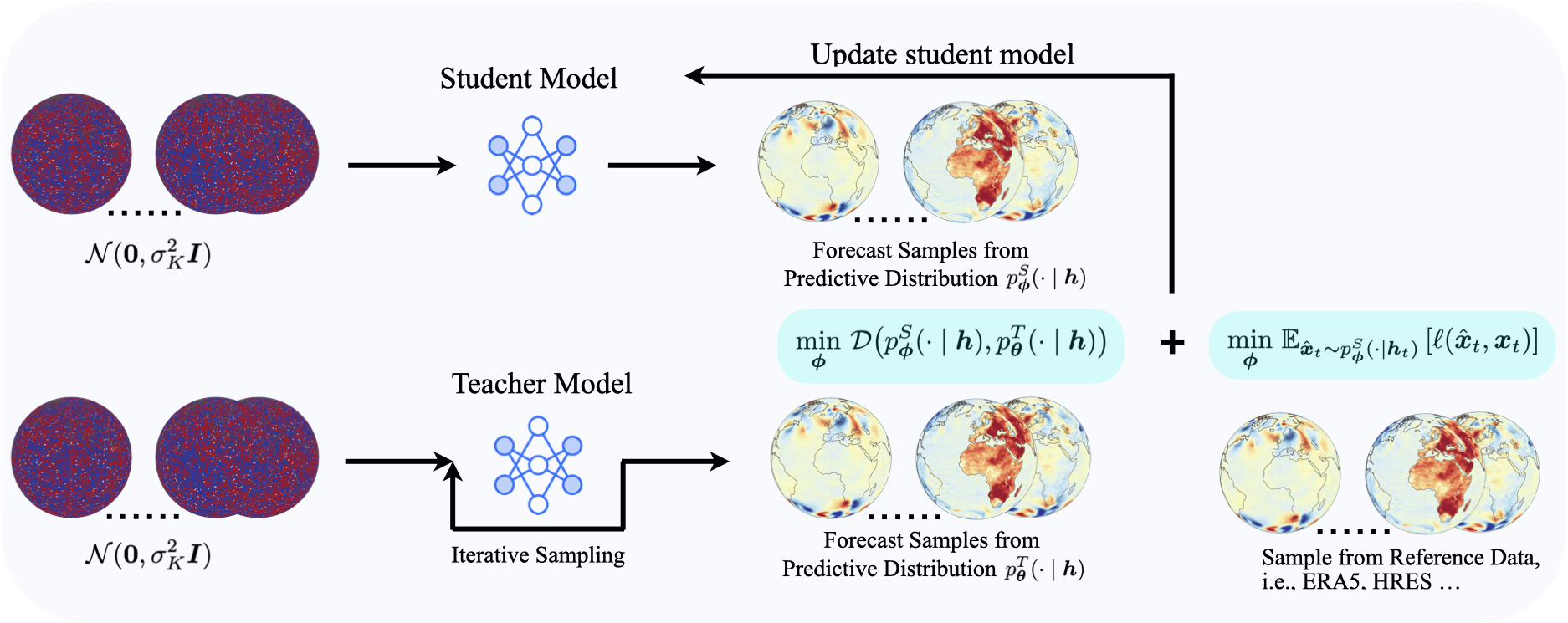}
    \caption{Overview of our distillation framework. Our objective aligns the single-pass student with teacher forecasts through distribution distance and improves accuracy using ground-truth supervision.}
    \label{fig:placeholder}
\end{figure}

\section{Background}
Let $\boldsymbol{x}_t\in\mathcal{X}\subset\mathbb{R}^d$ denote the ground-truth weather state at time $t$, and let $\boldsymbol{h}_t=(\boldsymbol{x}_{t-n},\ldots,\boldsymbol{x}_{t-1})$ denote its history. We omit time indices when clear and denote teacher and student forecast samples by $\boldsymbol{x}_{\mathrm{T}}$ and $\boldsymbol{x}_{\mathrm{S}}$, respectively.

\paragraph{Diffusion models and \textsc{GenCast}.} Diffusion models learn complex distributions by reversing a gradual noising process. \textsc{GenCast} \citep{price2025probabilistic} applies conditional diffusion to gridded atmospheric states and autoregressively generates global weather ensembles at 12-hour intervals.

For each forecast step, \textsc{GenCast} implements its conditional diffusion using the EDM \citep{karras2022elucidating}. During training, the future state \(\boldsymbol{x}\) is perturbed as
$
\boldsymbol{x}_k
=
\boldsymbol{x}
+
\sigma_k\boldsymbol{\epsilon}$, $
\boldsymbol{\epsilon}\sim\mathcal{N}(\boldsymbol{0},\boldsymbol{I}),
$
where \(\sigma_K>\cdots>\sigma_0\) defines the noise schedule. A denoiser $
D_{\boldsymbol{\theta}}
(\boldsymbol{x}_k,\sigma_k\mid\boldsymbol{h})
$
is trained to recover \(\boldsymbol{x}\). At inference, sampling starts from
$
\boldsymbol{x}_K
\sim
\mathcal{N}(\boldsymbol{0},\sigma_K^2\boldsymbol{I})
$
and iteratively denoises it to obtain forecast realizations. We denote the marginal at level \(k\) by \(p_{\boldsymbol{\theta},k}(\cdot\mid\boldsymbol{h})\) and the final predictive distribution by \(p_{\boldsymbol{\theta}}(\cdot\mid\boldsymbol{h})\). Further details are provided in Appendix~\ref{App: Diffusion Model - Gencast}.

\paragraph{Distribution Matching.}
Distribution matching learns a model distribution by minimizing its discrepancy from a target distribution. It is widely used in generative modeling, where the target is the data distribution, and in distillation, where the target is a pretrained teacher distribution \citep{yin2024one,yin2024improved}. For conditional weather forecasting, let \(p_{\boldsymbol{\phi}}^S(\cdot\mid\boldsymbol{h})\) and \(p_{\boldsymbol{\theta}}^T(\cdot\mid\boldsymbol{h})\) denote the student and teacher predictive distributions given historical states \(\boldsymbol{h}\). The general distribution-matching objective is
\begin{equation}
\label{eq:distribution-matching}
\min_{\boldsymbol{\phi}}\;
\mathbb{E}_{\boldsymbol{h}\sim p_{\mathrm{data}}}
\left[
\mathcal{D}\!\left(
p_{\boldsymbol{\phi}}^S(\cdot\mid\boldsymbol{h}),
p_{\boldsymbol{\theta}}^T(\cdot\mid\boldsymbol{h})
\right)
\right],
\end{equation}
where \(\mathcal{D}\) measures the discrepancy between two distributions. Popular instances of \(\mathcal{D}\) include KL divergence and Maximum Mean Discrepancy (MMD).

\section{Methodology}

\paragraph{Student generator.} Our student generator adopts the same architecture as the \(D_{\boldsymbol{\theta}}\) but generates each forecast in a single step. Given the historical states \(\boldsymbol{h}\) and a noise sample \(\boldsymbol{\epsilon}\), it produces a forecast as \(\hat{\boldsymbol{x}}=G_{\boldsymbol{\phi}}(\boldsymbol{h},\sigma_{K}\boldsymbol{\epsilon})\), where \(\boldsymbol{\epsilon}\sim\mathcal{N}(\boldsymbol{0},\boldsymbol{I})\). The induced predictive distribution is \(p_{\boldsymbol{\phi}}^S(\boldsymbol{x}_{t}|\boldsymbol{h}_t)=\int\delta\left(\boldsymbol{x}_{t}-G_{\boldsymbol{\phi}}(\boldsymbol{h}_t,\sigma_{K}\boldsymbol{\epsilon})\right)p(\boldsymbol{\epsilon})\mathrm{d}\boldsymbol{\epsilon}\) with the noisy marginal \(p_{\boldsymbol{\phi}}^S(\boldsymbol{x}|\boldsymbol{h})*\mathcal{N}(\boldsymbol{0},\sigma_k^2\boldsymbol{I})\).

\subsection{Supervised Distillation Objective}
Given the student distribution defined above, we formulate distillation using both teacher forecasts and ground-truth data. Standard distribution matching uses the teacher distribution as the sole target, causing the student to inherit both its forecast skill and errors. Since each historical state sequence \(\boldsymbol{h}\) is paired with a ground-truth future state \(\boldsymbol{x}\), we use this target as an additional supervised term. Our objective is defined as
\begin{equation}
    \label{eq: supervised distillation}
    \mathcal{L}(\boldsymbol{\phi})=\mathbb{E}_{\boldsymbol{h}_{t}\sim p_{\mathrm{data}}}\Big[\underbrace{\mathbb{E}_{\hat{\boldsymbol{x}}_{t}\sim p^S_{\boldsymbol{\phi}}(\cdot\mid \boldsymbol{h}_{t})}\left[\ell(\hat{\boldsymbol{x}}_{t},\boldsymbol{x}_{t})\right]}_{\mathcal{L}_{\mathrm{sup}}}+\beta \underbrace{\mathcal{D}(p^S_{\boldsymbol{\phi}}\left(\cdot|\boldsymbol{h}_t),p^T_{\boldsymbol{\theta}}(\cdot|\boldsymbol{h}_t)\right)}_{\mathcal{L}_{\mathrm{distill}}}\Big],
\end{equation}
where $\ell$ is a supervised loss, and \(\beta\) is the coefficient that balances the two terms. The second term measures the discrepancy between the student and teacher forecast distributions. 
\begin{wrapfigure}{r}{0.5\columnwidth}
\vspace{-2.3em}
\captionsetup{type=algorithm}
\vspace{1.5em}
\hrule height 0.9pt
\caption{Distillation training and inference}
\label{alg:distillation}
\hrule height 0.9pt
\begin{algorithmic}[1]
\footnotesize
\State \textbf{Training}
\State \textbf{Input:} dataset \(\mathcal{T}\), teacher \(D_{\boldsymbol{\theta}}\), sample count \(M\) and \(N\)
\State \(\boldsymbol{\phi}\gets\boldsymbol{\theta}\); freeze \(D_{\boldsymbol{\theta}}\)

\For{each training step}
    \State Sample \((\boldsymbol{x}_t,\boldsymbol{h}_t)\sim\mathcal{T}\)
    \State Draw \(M\) student samples
    \(\{\boldsymbol{x}_{t,S}^{(m)}\}_{m=1}^{M}\)
    
    \State Draw \(N\) teacher samples
    \(\{\boldsymbol{x}_{t,T}^{(n)}\}_{n=1}^{N}\)
    \State Compute \(\mathcal{L}(\boldsymbol{\phi})=\mathcal{L}_{\mathrm{sup}}(\boldsymbol{\phi})+\beta\mathcal{L}_{\mathrm{distill}}(\boldsymbol{\phi})\)
    \State
    \(\boldsymbol{\phi}
    \gets
    \boldsymbol{\phi}
    -\eta\nabla_{\boldsymbol{\phi}}\mathcal{L}\)
\EndFor
\Statex
\hrule 
\Statex \textbf{Inference}
\State \textbf{Input:}
\(\boldsymbol{h}_t\),
\(\boldsymbol{\epsilon}\sim
\mathcal{N}(\boldsymbol{0},\boldsymbol{I})\)
\State \textbf{Return:}
\(\hat{\boldsymbol{x}}_t
=
G_{\boldsymbol{\phi}}
(\boldsymbol{h}_t,\sigma_{\max}\boldsymbol{\epsilon})\)
\end{algorithmic}
\hrule height 0.9pt
\vspace{-2.2em}
\end{wrapfigure}
Here, it serves two purposes: distilling the distributional knowledge from the teacher model and regularizing training using the teacher’s predictive distribution as a reference prior. Since diffusion models define implicit distributions, standard density-based discrepancies such as KL divergence are intractable, while auxiliary density estimation can be unstable for high-dimensional weather data \citep{yin2024one,yin2024improved,zhou2024score,chen2025diffratio}. We therefore introduce a sample-based discrepancy derived from proper scoring rules.

\paragraph{Supervised loss \(\ell\).}
Supervising all ensemble members against the same observation may reduce diversity. In practice, we therefore use the ensemble-mean error
\(
\mathcal{L}_{\mathrm{sup}}
=
\left\|\bar{\boldsymbol{x}}_S-\boldsymbol{x}_t\right\|,\,
\bar{\boldsymbol{x}}_S
=
\frac{1}{M}\sum_{m=1}^{M}\boldsymbol{x}_S^{(m)},
\)
which improves mean accuracy without directly penalizing ensemble spread.

\subsection{Discrepancy from a Proper Scoring Rule}
To instantiate \(\mathcal{D}\) in Eq.~\eqref{eq: supervised distillation}, we use the energy score
\(
\mathrm{ES}(P,y)
=
\mathbb{E}_{X\sim P}\!\left[\|X-y\|\right]
-
\frac{1}{2}
\mathbb{E}_{X,X'\stackrel{\mathrm{iid}}{\sim}P}
\!\left[\|X-X'\|\right]
\)\footnote{Throughout, \(\|\cdot\|\) denotes the latitude- and variable-weighted norm used by \textsc{GenCast} \citep{price2025probabilistic}.},
a strictly proper scoring rule for multivariate distributions
\citep{szekely2004testing,gneiting2007strictly}. Minimizing its expected value under \(Q\) is equivalent to minimizing the squared energy distance
\begin{equation*}
\mathcal{D}_{\mathrm{E}}^2(P,Q)
=
2\mathbb{E}\!\left[\|X-Y\|\right]
-
\mathbb{E}\!\left[\|X-X'\|\right]
-
\mathbb{E}\!\left[\|Y-Y'\|\right],
\end{equation*}
where \(X,X'\stackrel{\mathrm{iid}}{\sim}P\) and
\(Y,Y'\stackrel{\mathrm{iid}}{\sim}Q\). Dropping the teacher-only term, the distillation objective becomes:
\(
\mathcal{L}_{\mathrm{distill}}
=
\mathbb{E}_{\boldsymbol{h}_t\sim p_{\mathrm{data}}}
\left[
\mathbb{E}\!\left[
\|\boldsymbol{X}_{\mathrm{T}}-\boldsymbol{X}_{\mathrm{S}}\|
\right]
-
\frac{1}{2}
\mathbb{E}\!\left[
\|\boldsymbol{X}_{\mathrm{S}}-\boldsymbol{X}'_{\mathrm{S}}\|
\right]
\right],
\)
where
\(\boldsymbol{X}_{\mathrm{T}}\sim
p_{\boldsymbol{\theta}}^T(\cdot\mid\boldsymbol{h}_t)\) and
\(\boldsymbol{X}_{\mathrm{S}},\boldsymbol{X}'_{\mathrm{S}}
\stackrel{\mathrm{iid}}{\sim}
p_{\boldsymbol{\phi}}^S(\cdot\mid\boldsymbol{h}_t)\). For each \(\boldsymbol{h}_t\), we draw \(M\) student samples
\(
\boldsymbol{x}_{\mathrm{S}}^{(m)}
=
G_{\boldsymbol{\phi}}
(\boldsymbol{h}_t,\sigma_{\max}\boldsymbol{\epsilon}^{(m)}),\) where \(\boldsymbol{\epsilon}^{(m)}
\sim\mathcal{N}(\boldsymbol{0},\boldsymbol{I}),
\)
and \(N\) teacher samples
\(\{\boldsymbol{x}_{\mathrm{T}}^{(n)}\}_{n=1}^{N}\)
through iterative sampling. An unbiased estimator
\citep{gretton2012kernel} is
\begin{equation}
\label{eq:u-statistics-distillation}
\widehat{\mathcal{L}}_{\mathrm{distill}}
=
\frac{1}{MN}
\sum_{m=1}^{M}\sum_{n=1}^{N}
\left\|
\boldsymbol{x}_{\mathrm{S}}^{(m)}
-
\boldsymbol{x}_{\mathrm{T}}^{(n)}
\right\|
-
\frac{1}{2M(M-1)}
\sum_{m=1}^{M}
\sum_{\substack{i=1\\i\neq m}}^{M}
\left\|
\boldsymbol{x}_{\mathrm{S}}^{(m)}
-
\boldsymbol{x}_{\mathrm{S}}^{(i)}
\right\|.
\end{equation}
Algorithm~\ref{alg:distillation} summarizes the training and inference procedures. Additional settings, including sample counts and loss weights, are provided in Appendix~\ref{subsubsec: Training details}.

\section{Experiment}
\label{section:Experiment}
We evaluate our method on global forecast skill, computational efficiency, and extreme events. We first compare its forecasting performance with the \textsc{GenCast} teacher and two distillation baselines, CM \citep{DBLP:conf/icml/SongD0S23} and DMD \citep{yin2024one}, at \(1^\circ\) resolution. We then assess extreme-event forecasting through typhoon-track prediction. For all methods, forecasts are generated with 25-member ensembles. Experimental settings and additional results are provided in Appendix~\ref{sec: experiments_settings}.

\begin{figure}
    \centering
    \includegraphics[width=1\linewidth]{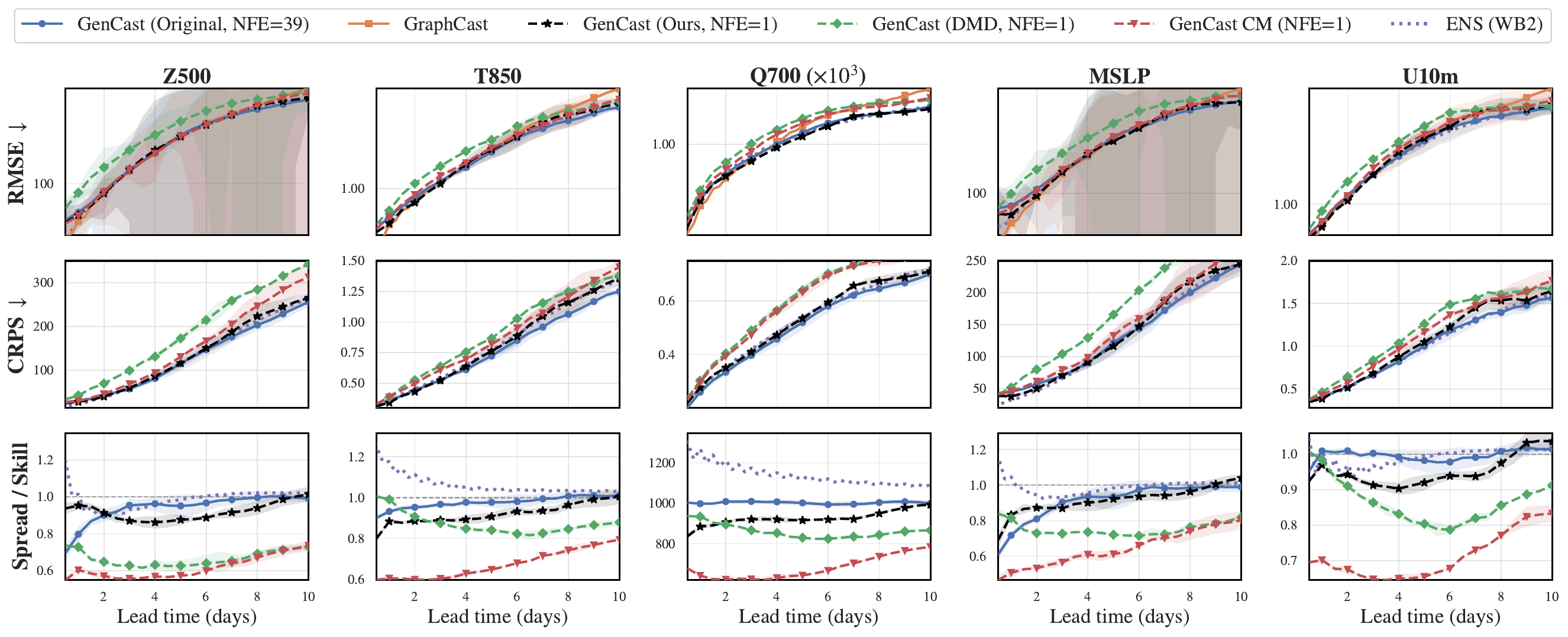}
    \caption{Global forecast skill across representative variables. Shading denotes \(\pm1\) standard deviation; lower RMSE/CRPS and a spread--skill ratio closer to one are better. ENS (WB2) denotes the 50-member ECMWF IFS ensemble.}
    \label{fig:distillation_metric}
\end{figure}

\paragraph{Global Weather Forecasting Skill.} We produce 10-day forecasts from 12 initial conditions, corresponding to the 1st and 15th of each month from July to December 2022. For reference, we also include \textsc{GraphCast} and the 50-member ECMWF IFS ensemble from WeatherBench 2 (ENS (WB2)); the latter is an operational physics-based NWP ensemble rather than an ML model \citep{rasp2024weatherbench2}. We use its ensemble mean for RMSE and the full ensemble for CRPS and spread--skill ratio. Figure~\ref{fig:distillation_metric} reports these metrics for five representative variables. Across variables and lead times, our student achieves the best RMSE and closely follows the teacher in CRPS among the distilled models. CM and DMD produce competitive ensemble means at short lead times, but their CRPS degrades more rapidly. Their spread-skill ratios indicate underdispersed ensembles, whereas our student better preserves the teacher’s calibration and forecast diversity.

\begin{figure}[t]
    \centering
    \includegraphics[width=1\linewidth]{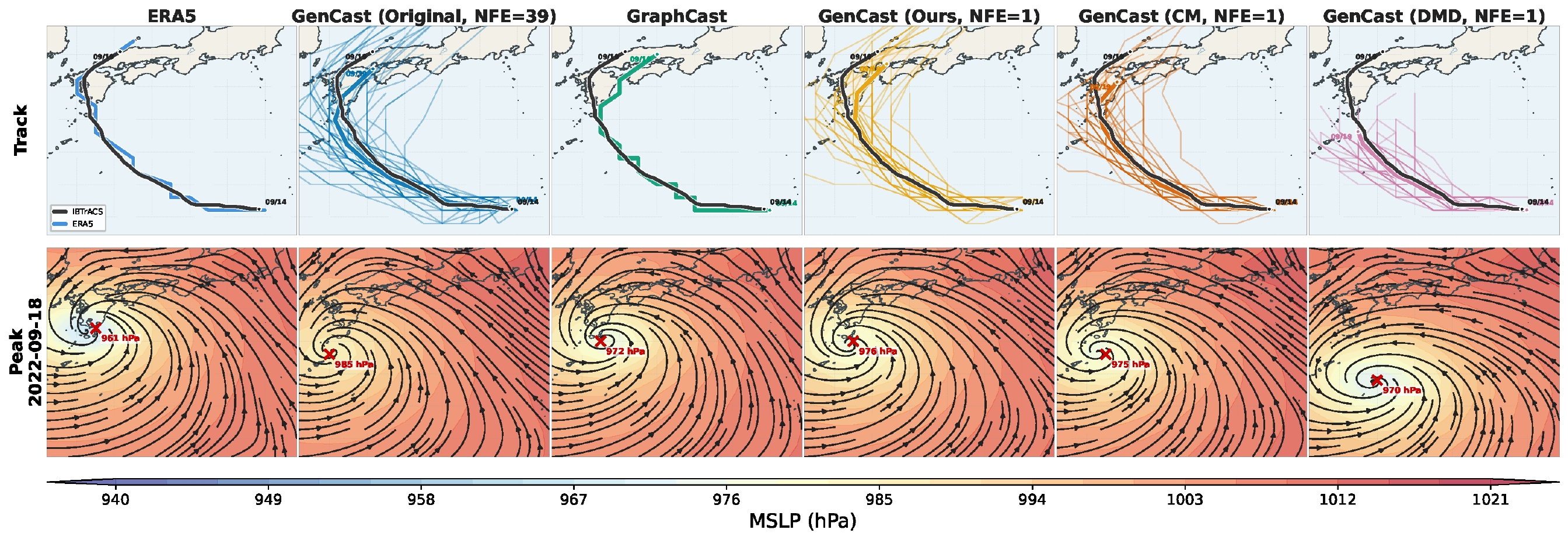}
    \caption{Typhoon Nanmadol track and peak-intensity forecasts. Top: observed and predicted tracks, with thin lines showing ensemble members and bold lines their means. Bottom: MSLP and 10-m winds at peak intensity, with crosses marking cyclone centers.}
    \label{fig:nanmadol_tracks}
\end{figure}

\paragraph{Extreme-Event Forecasting.}
We evaluate Typhoon Nanmadol over Japan using initial conditions from Sep 13, 2022, and autoregressively forecast its evolution through Sep 19. We derive tracks from each model's forecast fields using \textsc{TempestExtremes} \citep{ullrich2021tempestextremes} and compare them with the NOAA IBTrACS best track. As shown in Figure~\ref{fig:nanmadol_tracks}, our student preserves the storm trajectory and predicts a stronger cyclone core than the teacher. It also achieves the lowest position CRPS and deterministic track error among the probabilistic models, particularly at longer lead times (Figure \ref{fig:nanmadol_errors}, Appendix~\ref{fig:nanmadol_errors}).

\section{Conclusion}
\label{section:Conclusion}
We introduced a supervised distillation method for probabilistic weather forecasting. Our single-pass student outperforms CM and DMD while matching \textsc{GenCast} with 1 rather than 39 NFEs per autoregressive step. It also preserves ensemble spread and skillfully predicts Typhoon Nanmadol. Future work will explore more efficient distillation training and higher-resolution forecasting.



{\small
\bibliographystyle{plainnat}
\bibliography{ref}
}

\appendix
\clearpage
\section{Diffusion Model - \textsc{GenCast}}

\label{App: Diffusion Model - Gencast}
This section details the architectural and algorithmic formulations of \textsc{GenCast} \citep{price2025probabilistic}, which serves as the teacher model for our distillation framework. \textsc{GenCast} generates 15-day global ensemble weather forecasts with 12-hour temporal steps.

\paragraph{Diffusion Formulation.}
\textsc{GenCast} frames probabilistic weather forecasting as a conditional generative modeling task, factorizing the joint distribution of future atmospheric states autoregressively under a second-order Markov assumption: $p(\boldsymbol{x}_{2:T} \mid \boldsymbol{x}_1, \boldsymbol{x}_{0}) = \prod_{t=1}^{T-1} p(\boldsymbol{x}_{t+1} \mid \boldsymbol{x}_{t}, \boldsymbol{x}_{t-1})$. To stabilize learning dynamics across physical variables with vastly different numerical scales, the diffusion model does not generate the raw future atmospheric state $\boldsymbol{x}_{t+1}$ directly. Instead, it predicts a normalized residual state $\boldsymbol{z}_{t+1}$. The physical state is subsequently reconstructed via $\boldsymbol{x}_{t+1} = \boldsymbol{x}_{t} + \boldsymbol{s}\boldsymbol{z}_{t+1}$, where $\boldsymbol{s}$ is a diagonal scaling matrix that inverts the per-variable and per-level unit-variance normalization (with the exception of accumulated precipitation, which is predicted directly as $\boldsymbol{x}_{t+1} = \boldsymbol{s}\boldsymbol{z}_{t+1}$).

The continuous-time diffusion process largely adopts the Elucidated Diffusion Models (EDM) framework \citep{karras2022elucidating}. Importantly, to respect the $\mathbb{S}^2$ manifold of the Earth and avoid geometric distortions or pole singularities, GenCast departs from the standard practice of injecting i.i.d. Gaussian noise directly onto the latitude-longitude grid. Instead, it samples isotropic Gaussian white noise strictly on the sphere before projecting it onto the grid. Formally, at noise level \(\sigma\), the perturbed latent state is written as
\[
\boldsymbol{z}_{t+1}^\sigma=\boldsymbol{z}_{t+1}+\sigma\boldsymbol{\epsilon},\qquad\boldsymbol{\epsilon}\sim\mathcal{N}_{\mathbb{S}_2}\left(\boldsymbol{0},\boldsymbol{I}\right),
\]
where \(\mathcal{N}_{\mathbb{S}_2}\) denotes spherical isotropic Gaussian white noise \citep{lang2015isotropic}. A denoiser \(\boldsymbol{\epsilon}_{\boldsymbol{\theta}}\) is then trained to recover \(\boldsymbol{z}_{t+1}\) from \(\boldsymbol{z}_{t+1}^\sigma\), conditioned on the previous two atmospheric states \((\boldsymbol{x}_{t},\boldsymbol{x}_{t-1})\) and the noise level \(\sigma\). As in EDM, the model is parameterized to predict a denoised estimate under a noise-dependent preconditioning scheme, which improves optimization and sampling stability over a wide range of \(\sigma\), where  \(\sigma\in[0.03,80]\) in \textsc{GenCast}. Iterative sampling then progressively refines spherical noise into a coherent forecast realization, producing an ensemble approximation to the predictive distribution of future weather states. 

\paragraph{Denoiser Architecture.}
The core denoiser, $\boldsymbol{\epsilon}_{\boldsymbol{\theta}}(\boldsymbol{z}^{\sigma}_{t+1},\sigma\mid\boldsymbol{x}_{t}, \boldsymbol{x}_{t-1})$, utilizes the EDM preconditioning framework to dynamically scale the inputs, outputs, and skip connections based on the continuous noise level $\sigma$.

The model backbone $f_{\theta}$ handles the high-dimensional spatial resolution of the equiangular grid via an \textit{Encoder-Processor-Decoder} multi-graph architecture. The Encoder concatenates the noise-corrupted target $\boldsymbol{Z}_{\sigma}^t$ with the two autoregressive conditioning states along the channel dimension, mapping the dense grid data onto a highly compressed latent manifold: a 6-times refined icosahedral mesh comprising merely 41,162 nodes. Operating entirely within this reduced mesh space, the processor utilizes a specialized Spherical Graph Transformer. It consists of 16 consecutive transformer blocks (hidden dimension of 512, 4-head self-attention), where each node attends strictly to its 32-hop topological neighborhood on the mesh. To condition the network on the diffusion time step, the log-noise level $\sigma$ is embedded via sine-cosine Fourier features (32 frequencies) and a 2-layer MLP to obtain 16-dimensional encodings. These encodings modulate the activations through adaptive Conditional Layer Normalization (AdaLN). Finally, the decoder projects the processed mesh representations back to the dense physical grid.

\paragraph{Training Objective.} Unlike standard diffusion models that optimize a spatially uniform loss, \textsc{GenCast} \citep{price2025probabilistic} employs an objective tailored to both the spherical geometry of the Earth and the heterogeneous scales of atmospheric variables. Specifically, the preconditioned denoiser \(\boldsymbol{\epsilon}_{\boldsymbol{\theta}}\) is trained by minimizing a spatially and variable-weighted mean squared error over 40 years of historical reanalysis data ranging from 1979 to 2018 provided by the European Center for Medium-Range
Weather Forecasts (ECMWF) \citep{hersbach2020era5}:
\begin{equation}
    \label{eq:gencast_loss}
    \mathcal{L}_{\text{GenCast}}(\boldsymbol{\theta}) = \mathbb{E}_{t, \sigma, \boldsymbol{\epsilon}} \left[ \lambda(\sigma) \frac{1}{|G||J|} \sum_{i \in G} \sum_{j \in J} w_j a_i \left( \left[\boldsymbol{\epsilon}_{\boldsymbol{\theta}}(\boldsymbol{z}_{t+1}^\sigma, \sigma \mid \boldsymbol{x}_{t},\boldsymbol{x}_{t-1})\right]_{i,j} - z_{t+1, i,j} \right)^2 \right].
\end{equation}
Here, $G$ denotes the spatial grid indices, and $J$ denotes the diverse physical variables across multiple atmospheric pressure levels. $w_j$ provides static per-variable-level scaling weights to balance gradient magnitudes. Critically, $a_i$ introduces a latitude-dependent area weighting (normalized to unit mean) to physically counteract the severe geometric distortion and pole singularity inherent in the equiangular grid. $\lambda(\sigma)$ represents the continuous noise-level weight derived from the EDM formulation.

\paragraph{Sampler.}
At inference time, weather forecasts are generated by drawing an initial noise state $\boldsymbol{z}_{t+1}^{\sigma_{\max}} \sim \mathcal{N}(\boldsymbol{0},\sigma_{\max}^2\boldsymbol{I})$ and iteratively solving the reverse probability flow ordinary differential equation (PF-ODE) \citep{song2021scorebased,karras2022elucidating}. Under the variance-exploding formulation of EDM, this deterministic PF-ODE has the same marginal distributions as the stochastic process at each noise level $\sigma$:
\begin{equation}
    \mathrm{d}\boldsymbol{z}_{t+1}^\sigma = \frac{1}{\sigma}\left(\boldsymbol{z}_{t+1}^\sigma - \boldsymbol{\epsilon}_{\boldsymbol{\theta}}(\boldsymbol{z}_{t+1}^\sigma, \sigma \mid \boldsymbol{x}_{t},\boldsymbol{x}_{t-1})\right) \mathrm{d}\sigma.
    \label{eq:pfode_empirical}
\end{equation}
To generate a forecast, \textsc{GenCast} integrates Eq.~\eqref{eq:pfode_empirical} backward from $\sigma_{\max} = 80$ to $\sigma_{\min} = 0.03$ with a second-order numerical solver \textsc{DPM-Solver++} \citep{lu2025dpm} and stochastic churn. This integration uses $K=20$ solver steps and 39 NFEs per forecast step.

\section{Related Work}
\label{sec: related_work}
\paragraph{Learning Dynamical Systems.} Data-driven prediction of chaotic systems must balance short-term trajectory accuracy with long-term stability and statistical fidelity. Recent approaches use reservoir computing, neural operators, linear latent dynamics, and transformers for scalable long-horizon prediction \citep{pathak2018model,li2022learning,jiang2023training,cheng2025learning,he2025chaos,cheng2026information}. In parallel, learning-based data assimilation reconstructs dynamical states from partial observations \citep{bocquet2019data,fablet2021learning}. Latent-assimilation methods further reduce computational costs by performing updates in learned reduced spaces and bridging heterogeneous state and observation representations \citep{cheng2023generalised,cheng2024multidomain}, while recent systems extend learned assimilation to data-driven weather forecasting \citep{yang2025tensorvar,xu2025fuxida}. These directions improve learned dynamics and forecast initialization; our work addresses the complementary cost of generating probabilistic weather ensembles.

\paragraph{Probabilistic Weather Forecasting.} Recent weather models have increasingly moved toward probabilistic forecasting, with generative models playing a central role in representing uncertainties. For example, \textsc{SEEDS} \citep{li2024generative} and \textsc{GenCast} \citep{price2025probabilistic,cachay2025elucidated} use diffusion models to generate realistic, high-resolution stochastic weather ensembles, while \citet{cachay2024probablistic,cachay2025elucidated} extends diffusion forecasting to jointly generate multiple future forecasts. Although these models achieve strong probabilistic forecast skill, their iterative inference sampling remains computationally expensive. \citet{stock2025swift} proposes \textsc{Swift}, a one-step consistency model that accelerates diffusion models for efficient probabilistic forecasting. Newer architectures such as \textsc{FuXi-ENS} \citep{zhong2025fuxi}, \textsc{FourCastNet3} \citep{bonev2025fourcastnet}, and \textsc{WeatherNext2} \citep{alet2025skillful} achieve competitive one-step generation by producing ensembles in a single pass via learned stochastic perturbations. In parallel, hybrid models such as \textsc{NeuralGCM} \citep{kochkov2024neural} support ensemble forecasting through differentiable dynamics and learned physics parameterizations rather than explicit generative modeling.

\paragraph{Diffusion Distillation.} To improve inference efficiency, various distillation approaches have been proposed to compress multi-step sampling into one or a few steps. These approaches can be broadly categorized by how the student is aligned with the teacher: (1) \textit{Trajectory Matching Distillation} (TMD) and (2) \textit{Distribution Matching Distillation} (DMD). TMD methods, such as progressive distillation \citep{salimans2022progressive,meng2023distillation} and consistency models \citep{DBLP:conf/icml/SongD0S23,kim2024consistency,lu2025simplifying}, train the student to match the teacher’s trajectory or map intermediate states to the denoised endpoint. While effective, these methods are designed to preserve the consistency of the reverse diffusion process, which is indirect when the objective is to match the final predictive distribution, as in weather forecasting. Accordingly, DMD methods distill a teacher into a one-step generative model by minimizing the discrepancy between the final marginal distributions of the student and teacher. One popular choice of discrepancy is Kullback-Leibler (KL) divergence \citep{luo2023diff,wang2023prolificdreamer,yin2024one,xu2025one}; see \citet{wang2025uniinstruct} for a unified framework for divergences in diffusion distillation. However, for diffusion models, this KL divergence is generally intractable, and existing methods often require auxiliary networks to estimate either score functions \citep{yin2024one,yin2024improved,zhou2024score} or density ratios \citep{chen2025diffratio}. Recently, \citet{shao2025varflow} proposed a sample-based distillation method based on energy distance, followed by recent evidence of their effectiveness for training diffusion models \citep{pmlr-v267-de-bortoli25b}.

\section{Experiment Details}
\label{sec: experiments_settings}

\subsection{Experimental Setup}
\label{subsec: Experimental Setup}
\subsubsection{Data}
For all experiments, we use ERA5 reanalysis data \citep{hersbach2020era5} for atmospheric states and verification targets. We use \(1^\circ\) resolution, on which data are represented on a \(181\times360\) equiangular grid. We conservatively remap ERA5 fields from higher resolution to $1^\circ$ and use the six surface and six atmospheric variables at 13 pressure levels listed in Extended Data Table~1 of \citet{price2025probabilistic}. The student is trained on ERA5 data from January 1, 1979, through December 31, 2018, matching the \textsc{GenCast} training period, and evaluated on 12 initial dates: the 1st and 15th of each month from July to December 2022.

\subsubsection{Training details}
\label{subsubsec: Training details}
For the $1^\circ$ experiments, we train with a batch size of 4 for 20k AdamW steps on 4 H100 GPUs (96\,GB each). We use $M=N=24$ student and teacher samples. The loss-balancing coefficient is set to $\beta=2$, after rescaling the two loss terms to account for their different magnitudes.

All optimizers are implemented in Optax~\citep{deepmind2020jax}. We apply gradient clipping before the AdamW~\citep{loshchilov2018decoupled} update. The optimizer settings used for training the student models and the DMD critic network are identical and are summarized in Table~\ref{tab:student_opt}.

\begin{table}[h]
\centering
\caption{Optimizer hyperparameters.}
\label{tab:student_opt}
\begin{tabular}{ll}
\toprule
Hyperparameter        & Value \\
\midrule
Peak learning rate    & $5.0 \times 10^{-5}$ \\
Weight decay          & $0.01$ \\
Learning rate schedule & Cosine decay \\
Total training steps  & $20{,}000$ \\
EMA decay             & $0.999$ \\
\bottomrule
\end{tabular}
\end{table}

\paragraph{Tropical cyclone tracking.} 
To enable a fair comparison across forecasting models, tropical cyclone tracks are derived from the predicted meteorological fields of all models using a common external tracker, \textsc{TempestExtremes} \citep{ullrich2021tempestextremes}, rather than relying on model-specific tracking procedures. \textsc{TempestExtremes} is a general framework for feature detection and tracking on gridded atmospheric datasets and has been widely used for tropical cyclone analysis. Ground-truth tracks are taken from \textsc{IBTrACS} \citep{knapp2010international}, a global best-track archive that merges operational tropical cyclone records from multiple forecasting agencies. 

For each forecast, we apply the same detection criteria and temporal linking rules to all models. In particular, storm candidates are identified from the forecast fields at each lead time using a common set of cyclone-detection thresholds, and detections at consecutive lead times are linked into tracks using a fixed tracking configuration. The resulting forecast tracks are then matched to the corresponding \textsc{IBTrACS} best-track records for evaluation. This protocol ensures that differences in track skill reflect differences in the predicted meteorological fields rather than differences in post-processing or tracker design. 

\subsubsection{Metrics} Our evaluation uses area-weighted RMSE, CRPS, and the spread--skill ratio (SSR). Let $G$ be the set of spatial grid points and $w_i$ the normalized area weight at point $i$, with $w_i \propto \cos(\omega_i)$ and $\sum_{i\in G}w_i=1$, where $\omega_i$ is the latitude. Let $\hat{x}_t^{(m)}(i)$ be the $m$-th ensemble prediction of one variable at lead time $t$, and $x_t(i)$ the corresponding ERA5 target, with ensemble mean
\[
\bar{x}_t(i)=\frac{1}{M}\sum_{m=1}^M \hat{x}_t^{(m)}(i).
\]

\paragraph{RMSE} measures the magnitude of the forecast error and is used to assess deterministic accuracy. It is computed from the ensemble mean for probabilistic models and from the model prediction for deterministic models:
    \[
    \mathrm{RMSE}(t)=\sqrt{\sum_{i\in G} w_i \bigl(\bar{x}_t(i)-x_t(i)\bigr)^2 },
    \]

\paragraph{CRPS} measures probabilistic forecast accuracy by jointly assessing calibration and sharpness. For probabilistic models, we use the ensemble CRPS:
    \[
    \mathrm{CRPS}(t)=\sum_{i\in G} w_i
    \left[
    \frac{1}{M}\sum_{m=1}^{M}\bigl|\hat{x}_{t}^{(m)}(i)-x_t(i)\bigr|
    -\frac{1}{2M(M-1)}\sum_{m=1}^{M}\sum_{\substack{m'=1\\m'\neq m}}^{M}
    \bigl|\hat{x}_{t}^{(m)}(i)-\hat{x}_{t}^{(m')}(i)\bigr|
    \right].
    \]

\paragraph{Skill, spread, and SSR.}
Following the CRPS decomposition used for ensemble weather forecasts \citep{price2025probabilistic}, the two components measure complementary properties:
\[
\mathrm{Skill}(t)=\sum_{i\in G}w_i\frac{1}{M}\sum_{m=1}^{M}
\left|\hat{x}_{t}^{(m)}(i)-x_t(i)\right|,
\]
Here, \(\mathrm{Skill}\) measures member-wise error against the target, while
\[
\mathrm{Spread}(t)=\sum_{i\in G}w_i\frac{1}{M(M-1)}
\sum_{m=1}^{M}\sum_{\substack{m'=1\\m'\neq m}}^{M}
\left|\hat{x}_{t}^{(m)}(i)-\hat{x}_{t}^{(m')}(i)\right|
\]
\(\mathrm{Spread}\) measures ensemble diversity through pairwise member differences. Thus,
\(\mathrm{CRPS}=\mathrm{Skill}-\tfrac{1}{2}\mathrm{Spread}\) and
\(\mathrm{SSR}=\mathrm{Spread}/\mathrm{Skill}\). An SSR near one indicates calibrated dispersion; values below one indicate underdispersion, and values above one indicate overdispersion.

\subsection{Baseline Setup}

\paragraph{Consistency-model Distillation.}
To implement consistency distillation \citep{DBLP:conf/icml/SongD0S23} for \textsc{GenCast}, we treat the pretrained teacher as defining a family of conditional reverse trajectories in the normalized residual space, conditioned on the autoregressive history $\boldsymbol{h}=(\boldsymbol{x}_t,\boldsymbol{x}_{t-1})$ \citep{price2025probabilistic}. Since each forecast context provides only a single observed realization, we approximate the teacher conditional predictive distribution by drawing $m$ independent teacher samples. Concretely, for each training context, we initialize $\boldsymbol{z}_{\sigma_{K}} \sim \mathcal{N}(\boldsymbol{0},\sigma_{K}^2\boldsymbol{I})$ for $r=1,\dots,m$, and integrate the reverse PF-ODE along a discrete noise schedule $\sigma_K > \cdots > \sigma_2 > \sigma_1$. This yields $m$ paired intermediate states $(\boldsymbol{z}_{t+1}^{\sigma_{k+1},r}, \boldsymbol{z}_{t+1}^{\sigma_k,r})$ on matched teacher trajectories.

We then train a student network $f_{\phi}(\boldsymbol{z}_{t+1}^{\sigma},\sigma\mid\boldsymbol{x}_t,\boldsymbol{x}_{t-1})$, initialized from the pretrained \textsc{GenCast} denoiser, to produce predictions that are consistent across adjacent noise levels:
\[
\mathcal{L}_{\mathrm{CM}}(\phi)
=
\mathbb{E}_{t,k,r}
\left[
\left\|
f_{\phi}(\boldsymbol{z}_{t+1}^{\sigma_{k+1},r},\sigma_{k+1}\mid\boldsymbol{x}_t,\boldsymbol{x}_{t-1})
-
\mathrm{sg}\!\left(
f_{\phi}(\boldsymbol{z}_{t+1}^{\sigma_k,r},\sigma_k\mid\boldsymbol{x}_t,\boldsymbol{x}_{t-1})
\right)
\right\|_{\mathrm{GC}}^2
\right],
\]
where $\mathrm{sg}(\cdot)$ denotes stop-gradient, and $\|\cdot\|_{\mathrm{GC}}^2$ denotes the same latitude- and variable-weighted norm used in the \textsc{GenCast} training objective. At inference time, the student maps a single noisy input at $\sigma_{\max}$ directly to a forecast sample in one network evaluation, reducing sampling to a single-step generator.
\paragraph{Distribution Matching Distillation.} We implement DMD \citep{yin2024one} for \textsc{GenCast} using the same student generator \(G_{\phi}\), the frozen teacher denoiser \(D_{\theta}\), and an auxiliary denoiser \(D_{\psi}\) initialized from the teacher. Given a student forecast
\(\boldsymbol{x}_{S}=G_{\phi}(\boldsymbol{h},\sigma_{\max}\boldsymbol{\epsilon})\),
we perturb it as
\(\boldsymbol{x}_{S}^{\sigma}=\boldsymbol{x}_{S}+\sigma\boldsymbol{\epsilon}\).
The teacher and auxiliary denoisers estimate the teacher and student scores at \(\boldsymbol{x}_{S}^{\sigma}\), respectively. Their difference defines the distribution-matching objective
\[
\mathcal{L}_{\mathrm{DMD}}(\phi)
=
\mathbb{E}\!\left[
\frac{1}{2}
\left\|
\boldsymbol{x}_{S}
-
\mathrm{sg}\!\left(
\boldsymbol{x}_{S}
-
\frac{
D_{\psi}(\boldsymbol{x}_{S}^{\sigma},\sigma\mid\boldsymbol{h})
-
D_{\theta}(\boldsymbol{x}_{S}^{\sigma},\sigma\mid\boldsymbol{h})
}{\omega}
\right)
\right\|_{\mathrm{GC}}^{2}
\right],
\]
where \(\omega\) normalizes the score difference. The auxiliary denoiser is alternately trained on noisy student forecasts using the standard denoising objective. Following DMD, the generator is optimized with the distribution-matching loss and a regression term to teacher samples. At inference, the auxiliary denoiser is discarded, and \(G_{\phi}\) generates each forecast in one network evaluation. 

\subsection{Additional Experimental Results}
\begin{figure}[ht]
    \centering
    \includegraphics[width=0.7\linewidth]{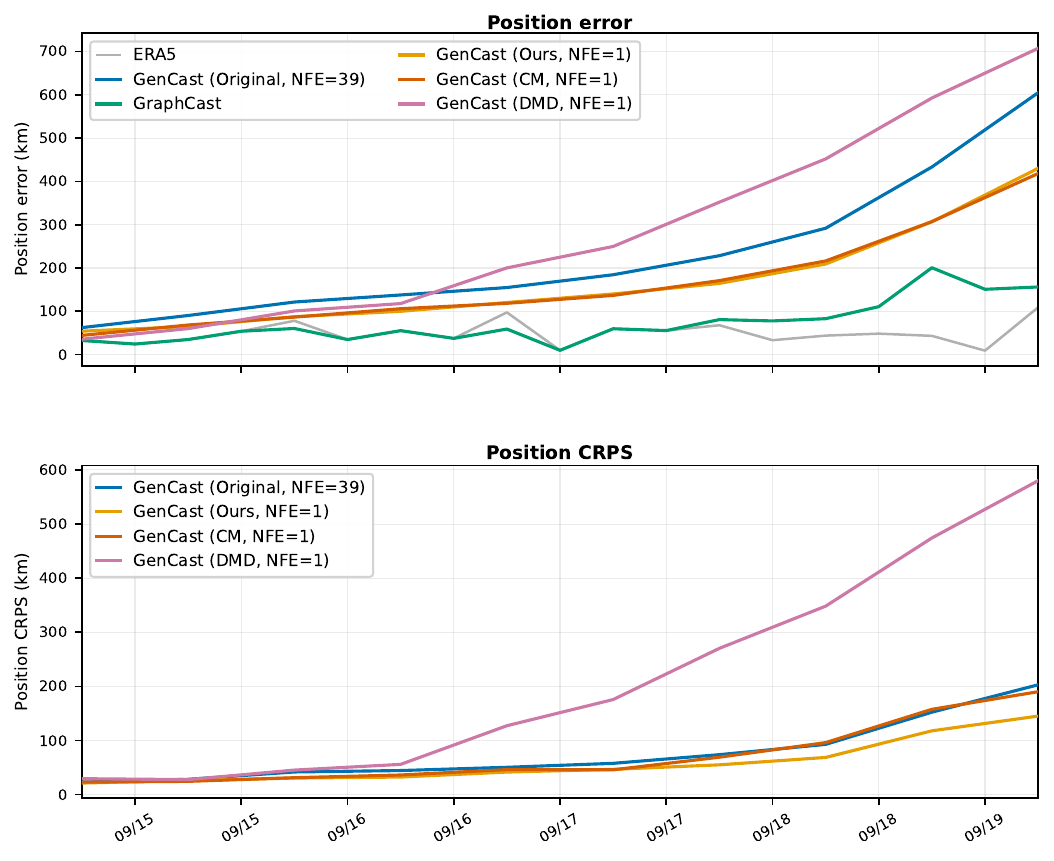}
    \caption{Typhoon Nanmadol track errors against the \textsc{IBTrACS} best track. Top: position error. Bottom: ensemble position CRPS, computed as the multivariate energy score over track locations.}
    \label{fig:nanmadol_errors}
\end{figure}

\begin{figure}
    \centering
    \includegraphics[width=0.75\linewidth]{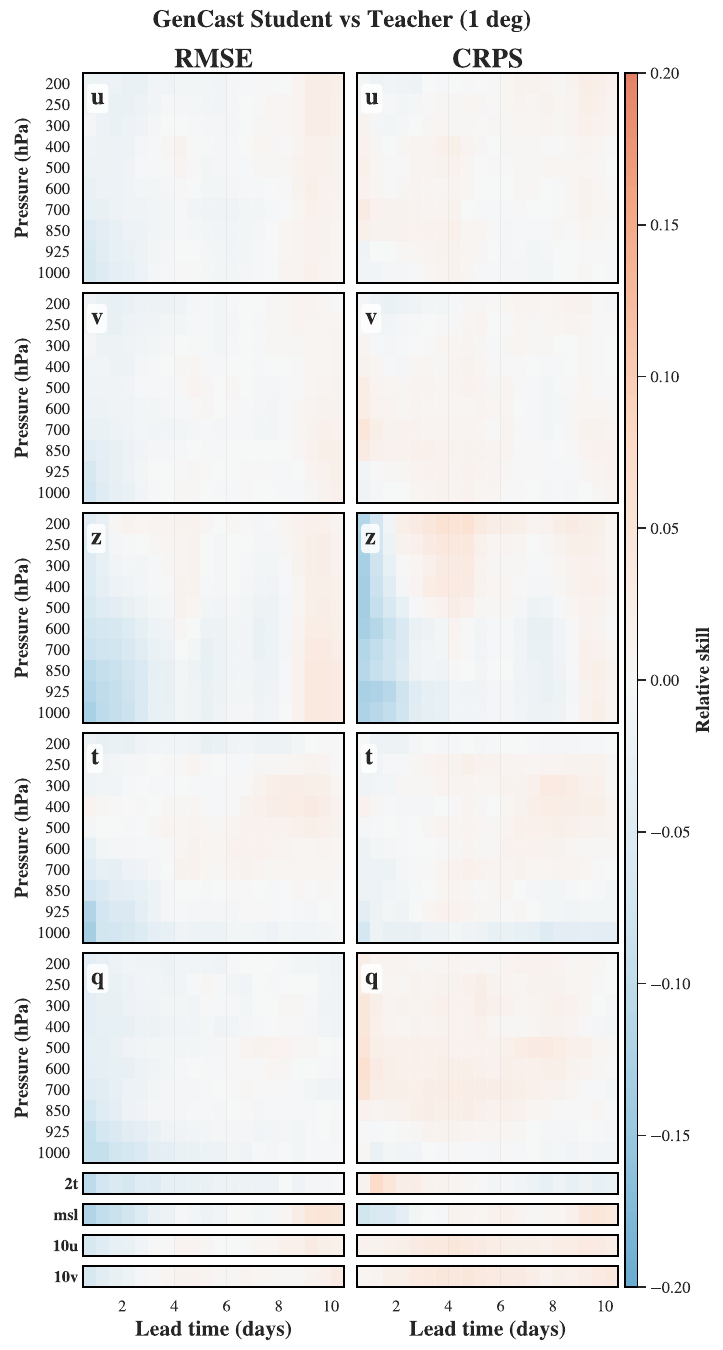}
    \caption{Relative RMSE and CRPS of our student compared with the \textsc{GenCast} teacher across variables, pressure levels, and lead times. Blue indicates improved skill and red indicates degradation.}
    \label{fig:scorecard}
\end{figure}

\paragraph{Full-variable comparison.}
Figure~\ref{fig:scorecard} extends the comparison across all surface variables and pressure levels. Our student matches or improves upon the teacher for most variables at short-to-medium lead times, with degradation concentrated at longer leads. This shows that the student broadly preserves forecast skill beyond the representative variables in the main text.
\paragraph{Typhoon-track errors.}
Figure~\ref{fig:nanmadol_errors} quantifies the Nanmadol results. Our student reduces the teacher's track-position error over most of the forecast and remains comparable to CM, whereas DMD diverges substantially after 16 September. More importantly, our student obtains the lowest position CRPS among the probabilistic models, indicating a better combination of track accuracy and ensemble dispersion.

\begin{figure}
    \centering
    \includegraphics[width=1\linewidth]{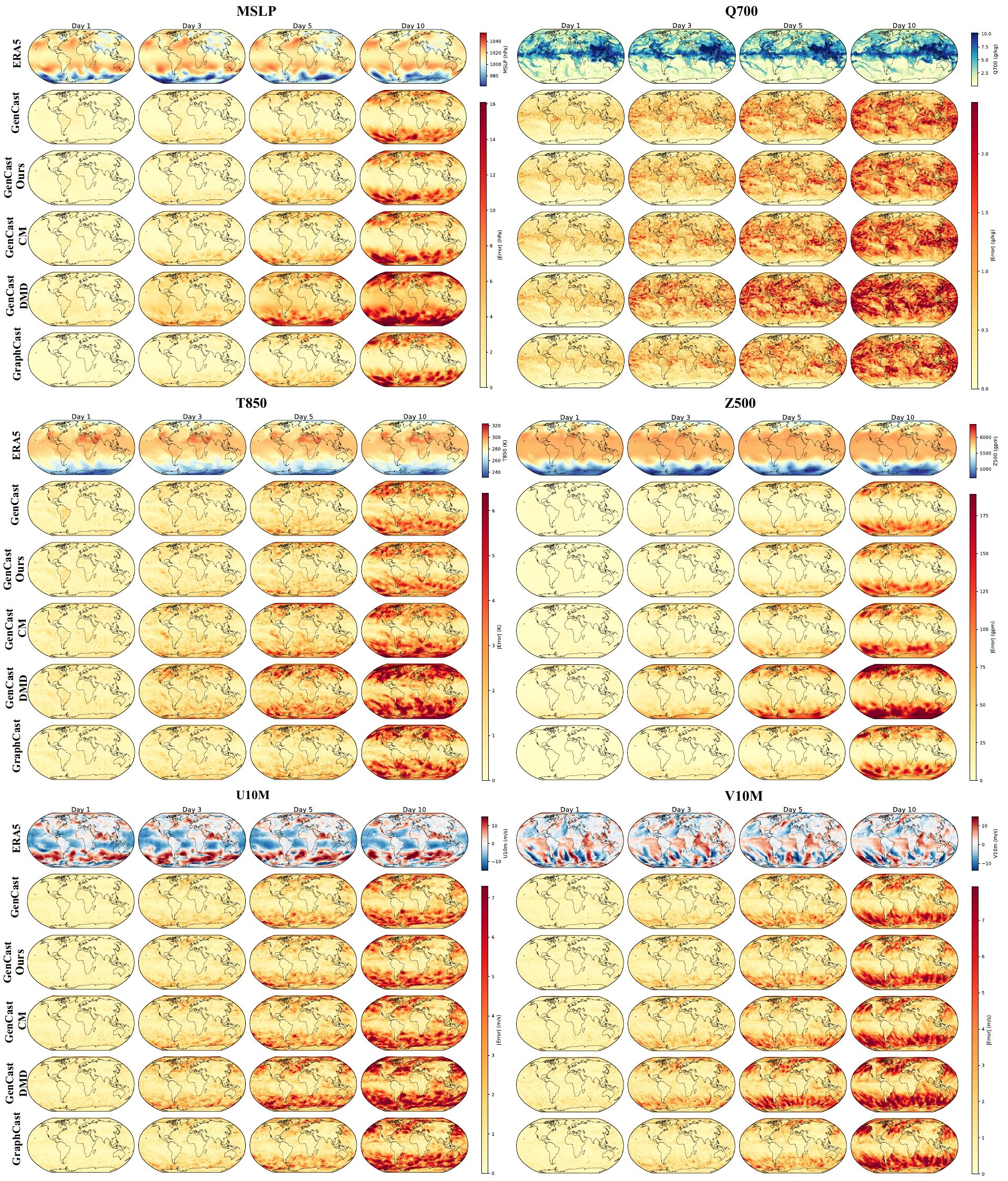}
    \caption{Spatial forecast errors for MSLP, Q700, T850, Z500, U10M, and V10M at lead times of 1, 3, 5, and 10 days, comparing the \textsc{GraphCast}, \textsc{GenCast} teacher, our student, CM, and DMD against ERA5.}
    \label{fig:spatial_errors}
\end{figure}

\begin{figure}
    \centering
    \includegraphics[width=1\linewidth]{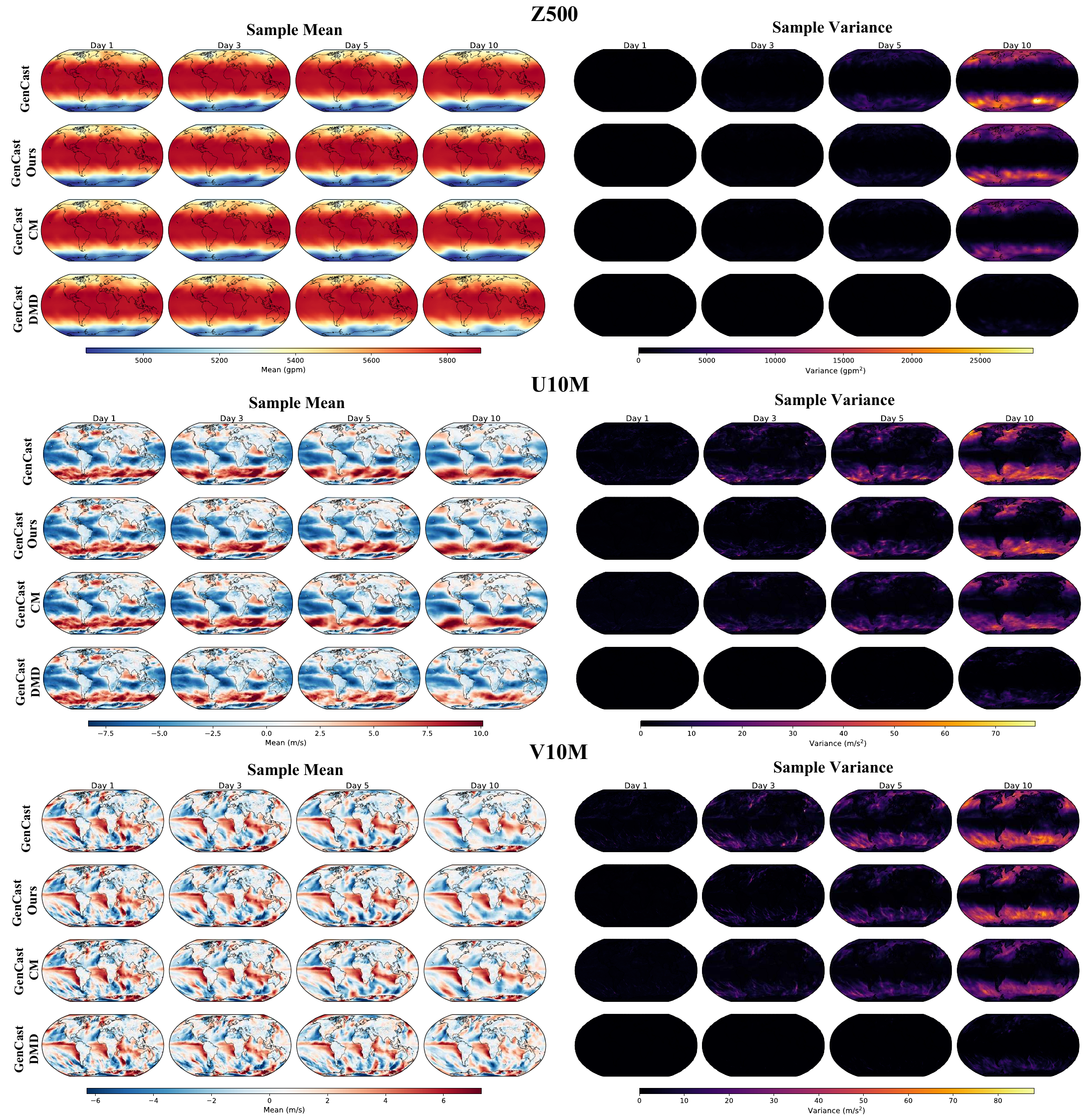}
    \caption{Spatial ensemble means and variances for Z500, U10M, and V10M. Columns show lead times of 1, 3, 5, and 10 days, and rows compare the multi-step \textsc{GenCast} teacher, our distilled student, CM, and DMD.}
    \label{fig:ensemble_statistics_zuv}
\end{figure}

\begin{figure}
    \centering
    \includegraphics[width=1\linewidth]{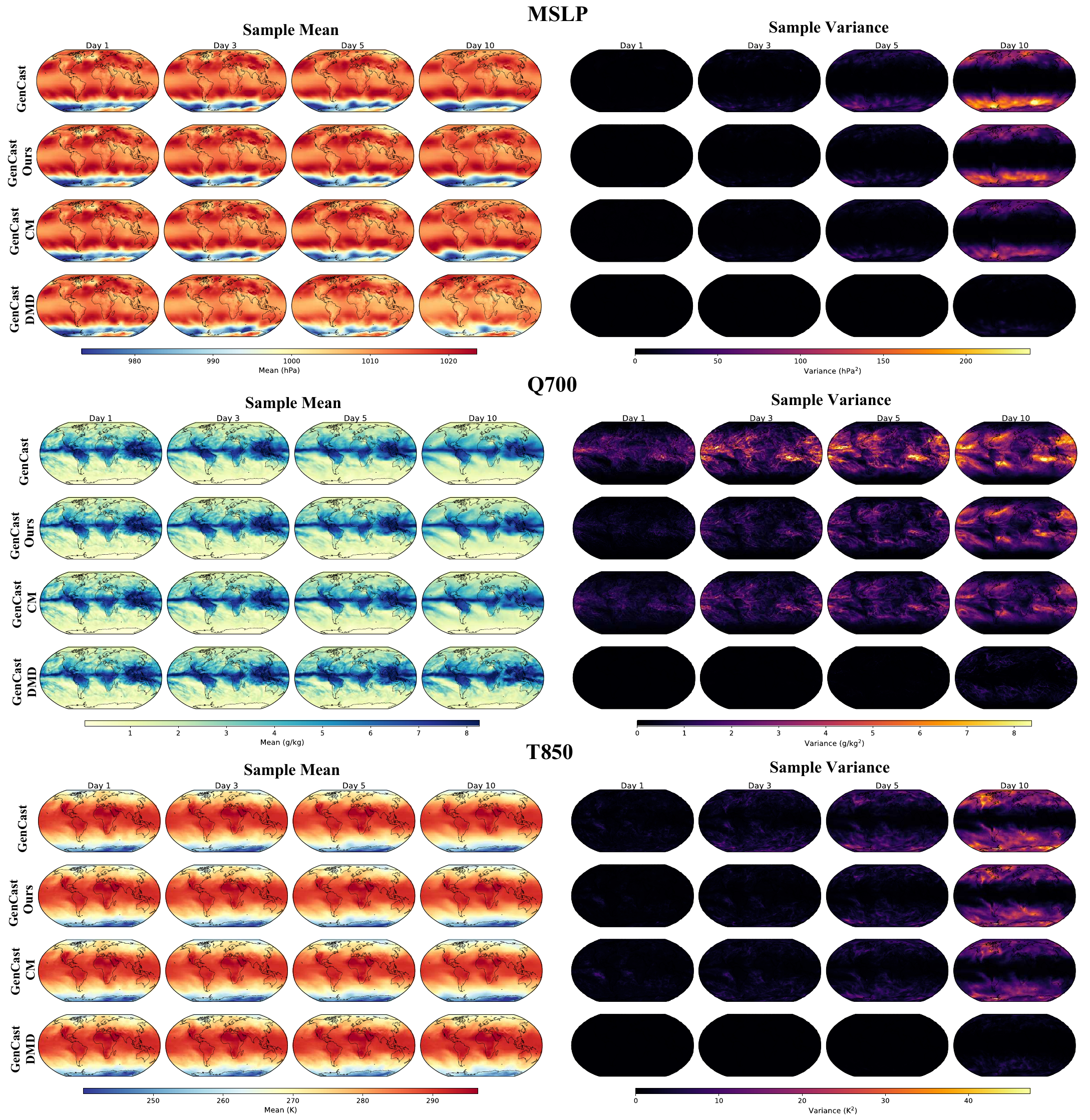}
    \caption{Spatial ensemble means and variances for MSLP, Q700, and T850. Columns show lead times of 1, 3, 5, and 10 days, and rows compare the multi-step \textsc{GenCast} teacher, our distilled student, CM, and DMD.}
    \label{fig:ensemble_statistics_mqt}
\end{figure}

\begin{table}[t]
\centering
\caption{Wall-clock time per autoregressive step on one NVIDIA GH200 with batch size 1. Values report mean \(\pm\) standard deviation over 10 runs, excluding JIT compilation, for the ensemble sizes shown.}
\label{tab:inference_time}
\small
\setlength{\tabcolsep}{10pt}
\renewcommand{\arraystretch}{1.15}
\begin{tabular}{lccc}
\toprule
  & GraphCast & GenCast (Original, NFE=39) & GenCast (Ours, NFE=1) \\
\midrule
Time (s) & $9.83\pm0.14$ &  $112.46\pm0.19$ & $21.44\pm0.16$ \\
Ensemble Size & 1 & 20 & 20 \\
\bottomrule
\end{tabular}
\end{table}


\end{document}